PAPER • OPEN ACCESS

# Automatic welding detection by an intelligent tool pipe inspection

To cite this article: C J Arizmendi *et al* 2015 *J. Phys.: Conf. Ser.* **628** 012082

View the article online for updates and enhancements.

# Automatic welding detection by an intelligent tool pipe inspection

**C J Arizmendi[1], W L Garcia [2,3], M A Quintero[3]**

[1] Mechatronic Department, Universidad Autónoma de Bucaramanga,
Avenue 42 N° 48-11, Bucaramanga, Colombia.
[2] System Engineering Department, Universidad Autónoma de Bucaramanga,
Avenue 42 N° 48-11, Bucaramanga, Colombia.
[3] Corrosion Research Institute, Km 2 via Refugio, Piedecuesta, Colombia.

**Abstract.** This work provide a model based on machine learning techniques in welds recognition, based on signals obtained through in-line inspection tool called "smart pig" in Oil and Gas pipelines . The model uses a signal noise reduction phase by means of pre-processing algorithms and attribute-selection techniques. The noise reduction techniques were selected after a literature review and testing with survey data. Subsequently, the model was trained using recognition and classification algorithms, specifically artificial neural networks and support vector machines. Finally, the trained model was validated with different data sets and the performance was measured with cross validation and ROC analysis. The results show that is possible to identify welding automatically with an efficiency between 90 and 98 percent.

## 1. Introduction

The pipelines are considered the most efficient way to carry fluids (oil and gas) over long distances [1]. Over time the integrity of this type of infrastructure is of concern for companies, mainly because much of the piping is nearing the end of its useful life due to the effect of external agents (corrosion, incorrect operations for civil works, landslides, vandalism etc.), which affects its operation. Consequently, it is necessary to monitor, evaluate and maintain pipelines to ensure reliability from a comprehensive analysis and reduce the probability of failure. The effects of a failure usually end up being accidents that impact the environment and population.

In recent decades, industry have developed different technologies to assess the condition of pipeline, of which one of the most used is the inspection with intelligent tools. An intelligent pipeline inspection tool[1]  (see figure 1) as defined in [2] is an instrumented vehicle is introduced into a pipeline travels through its extension, driven by the fluid itself and can perform various functions, such as:

- Measure pipeline operational condition.
- Detect and locate mechanical damages, defects and anomalies.

---

[1] Also known as ILI, In-line inspection, among others.

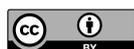

- Mapping the route of the pipeline, knowing the accurate position identify pipe movement and bending strain.

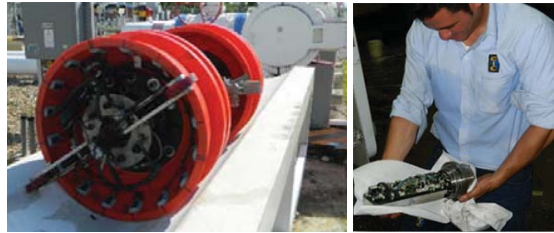

**Figure 1.** Smart tool inspection ITION and scraper.
**Source:** Ition-E project - CIC.

This type of intelligent tools has a number of sensors coupled to the internal and external of the chassis. The signals generated by the sensors are acquired, processed and stored while traveling through the pipe, then, large amount of data is downloaded, and artificial intelligence data analysis is applied to obtain relevant information, identification of discontinuities known as indications like anomaly, imperfection or component. Anomaly which are reported as possible segments of mechanical damage or pipe metal loss defects. Additionally, artificial intelligence detects piping accessories such as welds, coupling flanges, valves, shunts, among others.

Among the possibilities offered in-line inspection services, are implemented with different technologies depending on the purpose of the study of phenomena in pipes, vehicle technologies most commonly used are: MFL[2] [1], caliper [3], X-rays [4], ultrasound [5] and inertial [6]. From the point of view of data analysis, computational techniques used for processing, data analysis and recognition of welds are filtering signals and artificial intelligence.

Filtering techniques have been used in several works in order to improve the training data performance, for example in [1], which seeks to recognize defects in welds in pipes of carbon steel, based on information acquired by through a smart tool MFL inspection technology, tests were performed with Fourier [7] Wavelet [8] and Savitzky-Golay [9] filters to pre-process the input data seeking to eliminate irrelevant features that may hinder the classification defects. Validation results showed an improvement in the efficiency of the algorithm for almost all types of defects passing from 66.7% to 92.5%. Furthermore, it was able to confirm an improvement in training time using techniques such filtering. In [10], the wavelet transform is used to pre-process the data of 71 types of geometric defects in pipes, using data from a smart tool MFL inspection technology. Work results demonstrate that is possible with the aid of the Wavelet transform and other computational techniques to predict the depth and shape of these defects in 3D. In [11] - [12] Wavelet transform and a modified version of this transform is applied to remove noise in the data delivered by an intelligent tool MFL inspection technology to prevent recognition algorithms find defects wrongly. Finally, in [13], the wavelet transform is used to locate leaks in natural gas pipelines.

Different artificial intelligence techniques have been used for the recognition of phenomena in pipelines, in [1], [10], [14] neural networks were used to classify and identify some classes of geometrical defects in welds, in [15] using a genetic algorithm was possible to determine the shape, size and location of defects in about 1000 lines carrying natural gas. In [16] - [17], support vector machines applied to improve recognition performance of the algorithm, for defects in the metal loss and pipeline leaks.

---

[2] Magnetic Flux Leakage

In the present study, inspection data acquired by the smart tool to inspection trends integrity and operation, called ITION, developed by the Research Institute of Corrosion (CIC), for this work, inertial mapping technique is used. This purpose differs from other work (such as [1], [10], [11] and [16]) mainly because does not consist only of recognizing defects in the weld but also in locating it. In addition, inspection technology used by CIC is different (Inertial vs MFL), it is notable that no references about identification based on inertial welding techniques found.

Presented the prior differences, the aim of this work is to identify a weld using data acquired through an inspection tool with inertial technology from the study of computational intelligence applied to similar phenomena using other technologies inspection.

This paper is organized as follows. In section 2, the method is explained, describing each phase of the methodology and emphasizing the model developed. In section 3, the experiments performed together with the results obtained are shown, in section 4 discusses the results and section 5 the overall conclusions of the study.

**2. Methodology**
In order to achieve quality in the data mining process, the CRISP-DM methodology (Cross Industry Standard Process for Data Mining) [18] was applied. Next, main stages implemented are presented below:

*2.1. Business understanding*
First step was to understand the representative features of a weld. A weld defined in [19] as "fixed joining metal parts, made with or without melting the edges to be joined using a filler metal or not". In Oil and Gas transport a welding joint is a point or edge where two metal pipes or another component of the pipeline, valves, tees, flanges, branch, supports, are joined together by fusion bonding or metal to metal according to a particular geometry .An example of a weld is illustrated applied to join two tubes, figure 2.

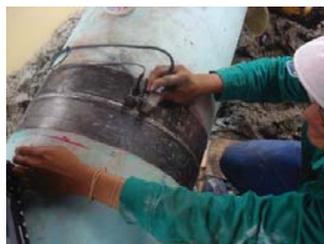

**Figure 2.** Welding pipe applied in section.
**Source:** Ition-E project - CIC.

When the intelligent line inspection passes by a welding, weld acts as a barrier and attempts to stop its path due to the excess material that is adhered to the inner surface protruding tube, however, at that moment and thrust forces phenomena occurs of inertia about the tool which keeps the dynamics and certain sensors recorded this tool behaviour.

Find all pipeline welds is an important purpose for the operator. During construction of the pipeline, manufacturer supplies the location of all welds, however, over time, repairs or changes occur due to maintenance which in some cases are not documented or reported, which generates the obsolescence of this information. Additionally, vandalism install accessories as valves in order to extract illegally pipeline products, in this case, apply welding to join the valve to the pipe. By construction conditions, the average distance between welds not exceeding twelve (12) meters. The objective of the analysis of welds from variables defined by experts is to identify sites where the welds are located with less than ten (10) percent error.

*2.2. Data understanding*
According expert criteria to identify a welding, information input variables shown in Table 1, is required

Table 1. Variables used by the experts to identify a weld

| Variable | Notation | Unit | Accepted value | |
|---|---|---|---|---|
| | | | Maximum | Minimum |
| Rotation | Gh | lsb | 6000 | -6000 |
| Acceleration | Ah | lsb | 5454 | -5454 |
| Magnetization | Mh | lsb | +2500 | -2500 |
| Vibration | Vh | lsb | 32000 | -32000 |
| Position | Od | m | 50000 | 0 |

Regarding the performance experienced by the tool during its journey through the pipe the three axes of rotation is recorded, which are variables that measures changes in the angular velocity, acceleration also recorded in all three axes, measuring the speed changes linear magnetization measures changes in the magnetic field remaining in the pipe due to the previous step of a magnetic arrangement. The vibration sensor records changes in the intensity of the mechanical vibrations generated by the friction between the tool body and the pipe walls. The position reference variable shows the axial distance (i.e. parallel to the flow direction) to measure the length of travel of the tool through the pipe.

The values delivered by the rotation, acceleration and magnetization are represented in the spatial axes (X, Y, Z), that is, there is data for each axis. In total have 12 input variables, 3 of acceleration, 3 of rotation 3, 3 related to the variation of the magnetic field, 2 of vibration and position. All data are numerical and are in the same unit related to the resolution of digital analog converter called *lsb* except the sensor position, which is measured in meters.

It was requested the experts that be defined an output variable, which welding was called, in the table 2 is shown the characteristics of this variable.

Table 2. Characteristic of the variable output

| Variable | Notation | Data type | Accepted value | |
|---|---|---|---|---|
| Weld | Weld | Text | S | N |

Then, data were reviewed to discard values outside the permitted range, based on information provided by experts (see table I). An exploration of a given set of data was performed and found the following assumptions:

- When the tool passes a weld, vibration signals show a higher intensity than in most other pipe segments.
- The acceleration magnitude, this being understood as the square root of the sum of the X, Y, Z squared is less than or equal to the magnitude of the acceleration before reaching the welding possible. This proves that the tool is "slowing" in its initial phase of step welding.

- The magnitude of the acceleration is greater than or equal after going through the possible welding. This means that tool speeds for an instant of time when the solder is overcome.

Finally, developed and implemented an algorithm, which was applied in all data in order to eliminate those records with values outside the permitted ranges for each variable or was unrepresentative values. After performing this task, is determined for example, using only the information recorded by the tool once was moving through the pipe. This decision was made because it was found that when the tool is installed at the point of release, spends a considerable period of time before run into the pipeline. While the tool is stationed at the launch site captures information that is not representative for the study of the inertial behaviour welds.

*2.3. Data Preparation*

The information provided was delivered in a text file in comma separated csv extension columns. The first row of the file represents the names of each of the variables, both inlet and outlet. The other lines contain the data for each variable. Then, in order twenty constructed data files, the former contains 2.5% of welds and twentieth file contains 50% of the welds present during inspection. That is, each file contains about 2.5% more than the previous welds. However, the results of this work will be presented based on four of the twenty files built, with the following number of records, hereafter levels: 148, 642, 1464 and 1838 to be used in the training phase and a file 600 instances as test data. Each data set contains the same number of instances classified as the output variable as welding and soldering, for example, in the case of assembly 148, there are instances with the solder 74 and 74 instances label as welding. The output variable was validated considering information from a document called Pipe Tally Data by the owner of the pipeline, which accurately describe the location of welds applied during construction.

The 5 files are converted to csv format to format arff through the Weka tool. Weka is one of the most comprehensive data analysis software, using different pre-processing techniques and data modeling tools that implements. Developed with Java technology which allows it to run on almost any platform, is freely available under GNU license and its libraries can be easily integrated into other software applications.

Taking advantage of the benefits offered this software tool and following the recommendation of the work reported in the literature, applies preprocessing to reduce the number of input variables. The attribute selection technique and algorithm CfsSubsetEval [20] is used. In total, 10 files are created in the initial five arff file type with all the input variables and the other five products pre-processing.

*2.4. Modeling*

In figure 3 the proposed model for the recognition of welds, which consists of two phases, the first called learning is part of the generated files in the pre-processing to train a classification algorithm, which according to illustrated the work reported in the literature may be artificial neural networks [1] or support vector machines [16]. Then, in the evaluation phase, each classification algorithm is evaluated using cross validation technique [21] and a set of tests not used during training. The results of the evaluation are validated by two metrics, the percentage error and the ROC [22] analysis. Whether validation results meet the stated objective process, then process ends, otherwise, it returns to the learning phase to review the causes of the problem and rethink a new learning scenario with the new settings.

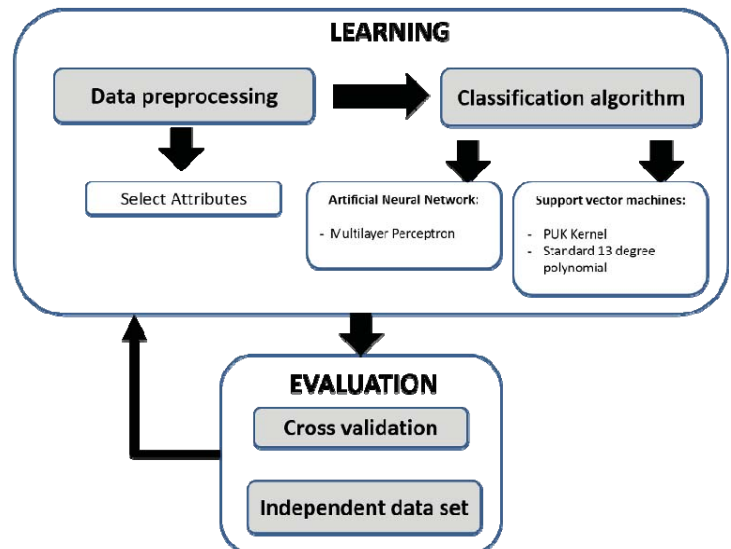

**Figure 3.** Model used for recognition of welding

**3. Experiments and results**
The information which the experiments were developed corresponds to an intelligent inspection in a Colombian pipeline in September 2012, with the following specifications:

• 12" nominal diameter.
• 36 km inspection.
• 2'150.835 records with an approximate size of 3GB.
• 3000 welds.

As it were mentioned in section 2 using the Weka software, 10 dataset was constructed, half of them are pre-processed using the technique of selecting attributes and the other half were left with all the input variables defined by the expert. The data for training were taken from a section of the inspection and consecutively, each training file has the same number of records with no welding and welding. Then, these files were used during training model for learning an artificial neural network of the multilayer perceptron and support vector machine with kernel called PUK or Pearson Universal Kernel5 [23] function. The performance of each model is evaluated using cross validation and independent set of data with a total of 600 instances, which, were taken from the final section of the inspection. It is taken into account as a criterion of success that the error rate with two evaluation metrics is less than ten percent. For each of the tests were performed to model eight (8) set of instances experiments, four (4) of the eight (8) experiments twelve (12) defined by the input variables used and the expert is mentioned in the understanding phase of data in section 2. In the other four (4) experiments, a feature selection algorithm, which recommended the use of certain input variables for each case was used.

*Model Tests 1838 instances*

For tests with this data set, in four (4) of the eight (8) experiments, CfsSubsetEval algorithm recommended using Y-axis acceleration, Y-axis magnetization with two vibration was employed. According to the results obtained when evaluating each model against a set of 600 instances untrained and cross-validation by the metric error rate (see table 3). Shows that two models recognize the location of the weld clearly when the variables acceleration (AHY) magnetic field in the pipe axis (MHY) and vibration (VH2) are used. The trained model that best meets the evaluation data is to support vector machines with PUK kernel with less than two (2%) percent error.

TABLE 3 Error rate models using 1838 instances in training

| Pre-processing technique | Artificial neural networks | | Support vector machines | |
|---|---|---|---|---|
| | Cross validation 10 folds | Independent set 600 instances | Cross validation 10 folds | Independent set 600 instances |
| None | 2.67 | 12.19 | 1.41 | 7.33 |
| Select attributes (Ahy, Mhy, Vh2) | 3.67 | 4.98 | 2.61 | 1.33 |

Results of the tests conducted in set of 600 instances and cross-validation metrics, but using ROC curves (see Table 4) shows that when pre-processing is applied, the false positive rate (number of records evaluated where the algorithm obtained recognition as a false output and was actually true) and true negative (number of records evaluated where recognition algorithm obtained as true output but is actually false) decreases compared with models that do not use the pre-processing. Furthermore the model that implements an algorithm of support vector machines is more accurate than the model with the algorithm of artificial neural networks (the value of the sensitivity and specificity is closer to 100% compared to the model that implements the artificial neural network).

TABLE 4 ROC analysis using models 1838 instances in training

| Pre-processing technique | Artificial neural networks | | | | Support vector machines | | | |
|---|---|---|---|---|---|---|---|---|
| | Cross validation 10 folds | | Independent set 600 instances | | Cross validation 10 folds | | Independent set 600 instances | |
| | TPR | SPC | TPR | SPC | TPR | SPC | TPR | SPC |
| None | 99.2% | 99.0% | 99.7% | 89.3% | 99.2% | 99.3% | 99.7% | 93.0% |
| Select attributes (Ahy, Mhy, Vh2) | 98.6% | 98.9% | 100% | 98.0% | 98.3% | 99.1% | 99.7% | 99.0% |

**4. Discussion**
With the results, it can be concluded that, using only 3 of the 12 input variables, the two developed models recognize the weld. This benefit by reducing processing time and analysis, considerably as large amount of data is processed.

The attributes that offered better performance for the developed models were acceleration and magnetic field variation on the Y axis and one of the vibration sensors. The pipeline inspection tool is configured such that the axial axis is the axis Y. This is consistent with a priori the experts have identified. However, during testing, as the number of data is increased, the rotation information in all three axes was not classified by the algorithm attribute selection as a representative and was in these assemblies where the best results were obtained model. This finding contradicts the opinion of experts who consider this information during the analysis process. The reason why the model uses only one of the vibration sensors is because one of them is filtered and real time processing which phenomenon intensifies the welding step and reduces the dynamic noise.

Regarding related work in the literature, a similar procedure to that reported in [1] and [16], was employed (i.e., pre-processing, training based on artificial intelligence and data validation) , the results demonstrate the feasibility of use of these computational techniques for identifying welds.

**5. Conclusions**
The results of this work show that welds can be identified automatically using artificial intelligence techniques and data mining with a percentage error of less than five percent (5%). Furthermore, it is concluded that 3 of 12 input information variables is only necessary to locate a weld: the acceleration and change of magnetic field in the Y axis and the vibration sensors.

It can be concluded that, under the conditions established by each test, it is possible to automatically recognize the position where the welds are located in a pipeline. The foregoing allows experts now have a versatile tool to provide timely reports. . However, as the developed models have a margin of error, in such cases, is at the discretion to the expert determine whether the point has or not welding, according to their experience and pipeline design information.